\documentclass[conference]{IEEEtran}
 \pdfoutput=1
\usepackage{graphicx,epsfig}
\usepackage{amsmath,amsfonts,amssymb}
\usepackage[ruled]{algorithm}
\usepackage{algpseudocode}
\usepackage{url}

\newcommand{\xb}{\mathbf{x}}
\newcommand{\xbt}{\tilde{\mathbf{x}}}
\newcommand{\yb}{\mathbf{y}}
\newcommand{\ybt}{\tilde{\mathbf{y}}}
\newcommand{\zb}{\mathbf{z}}
\newcommand{\zbt}{\tilde{\mathbf{z}}}

\newcommand{\zetaB}{\mbox{\boldmath$\zeta$}}

\begin{document}
%
\title{Performance Evaluation of Random Set Based Pedestrian Tracking Algorithms}

\author{\IEEEauthorblockN{Branko Ristic}
\IEEEauthorblockA{
ISR Division\\
DSTO\\
Australia\\
{branko.ristic@dsto.defence.gov.au}} \and \IEEEauthorblockN{Jamie
Sherrah }
\IEEEauthorblockA{ISR Division\\
DSTO\\
Australia\\
jamie.sherrah@dsto.defence.gov.au} \and \IEEEauthorblockN{\'{A}ngel
F. Garc\'{\i}a-Fern\'{a}ndez }
\IEEEauthorblockA{Department of Signals and Systems\\
Chalmers University of Technology\\
Sweden \\
angelg@chalmers.se}
%
}

\maketitle

\begin{abstract}
The paper evaluates the error performance of three random finite set
based multi-object trackers in the context of pedestrian video
tracking. The evaluation is carried out using a publicly available
video dataset of 4500 frames (town centre street) for which the
ground truth is available. The input to all pedestrian tracking
algorithms is an identical set of head and body detections, obtained
using the Histogram of Oriented Gradients (HOG) detector. The
tracking error is measured using the recently proposed OSPA metric
for tracks, adopted as the only known mathematically rigorous metric
for measuring the distance between two sets of tracks. A comparative
analysis is presented under various conditions.
\end{abstract}

%
\IEEEpeerreviewmaketitle

\section{Introduction}

Random set theory has recently been proposed as a mathematically
elegant framework for Bayesian multi-object filtering
\cite{mahler_07}. Research within this theoretical framework has
resulted in new multi-object filtering algorithms, such as the
probability density hypothesis (PHD) filter \cite{mahler_phdf_03},
Cardinalised PHD filter \cite{mahler_cphdf_07}, and Multi-Bernoulli
filter \cite{mahler_07, vo_vo_cantoni_09}. The main feature of
multi-object filters is that they estimate sequentially the number
of objects in the surveillance volume of interest (the so-called
cardinality) and their states in the state space. Formulation of
multi-object {\em trackers} from random-set based multi-object {\em
filters} has attracted a lot of interest recently, see e.g.
\cite{clark_sonar,pham_07,maggio_08,sherrah_11}. The output of a
tracker is a set of {\em tracks}, that is, labeled temporal
sequences of state estimates associated with the same object.
\begin{figure*}[htbp]
\vspace{-7cm}\centerline{\includegraphics[width=16cm]{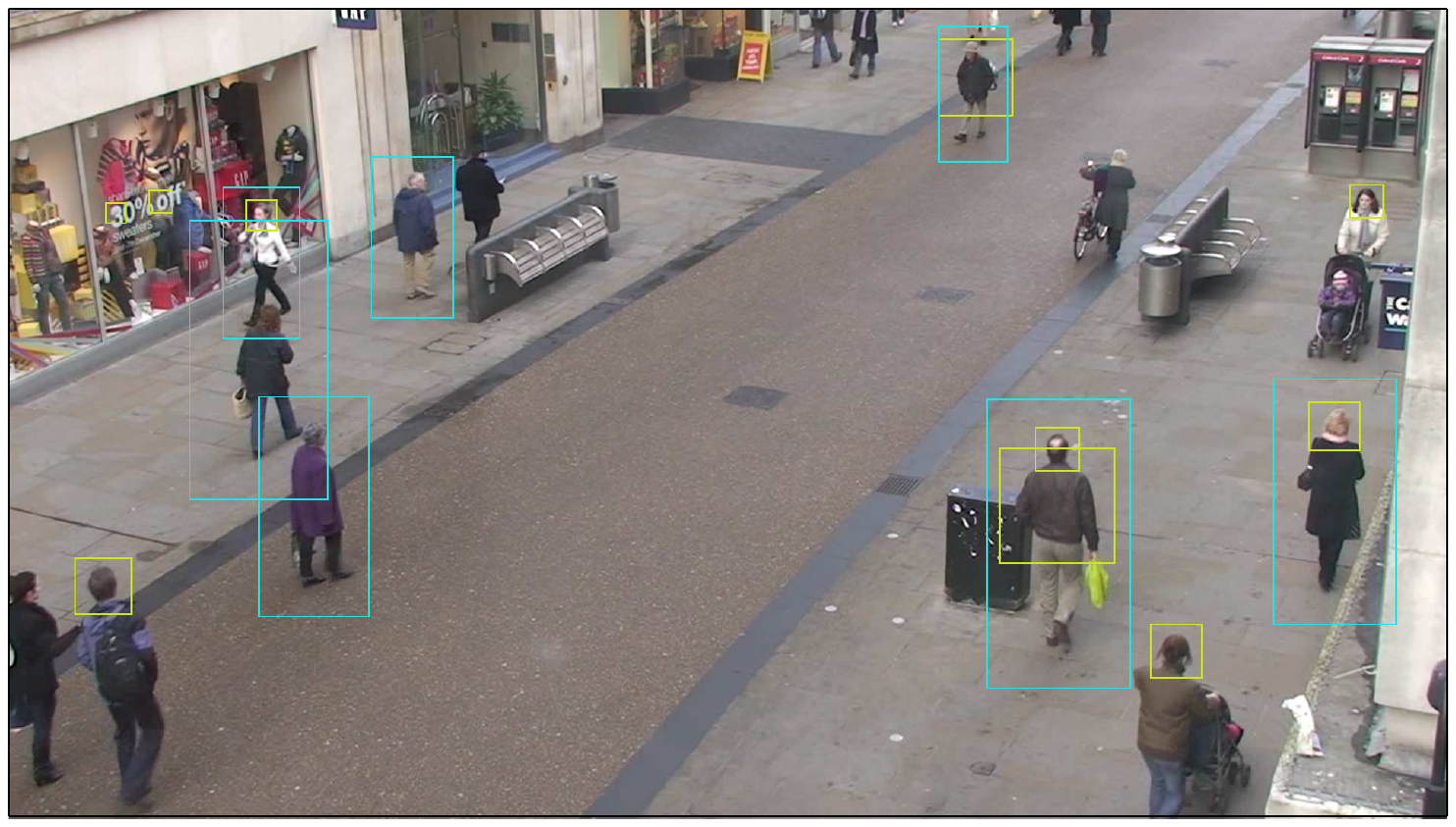}}
\vspace{-8cm}\centerline{\footnotesize (a)\vspace{-7cm}}
\mbox{.}\vspace{-2cm}
\centerline{\includegraphics[width=16cm]{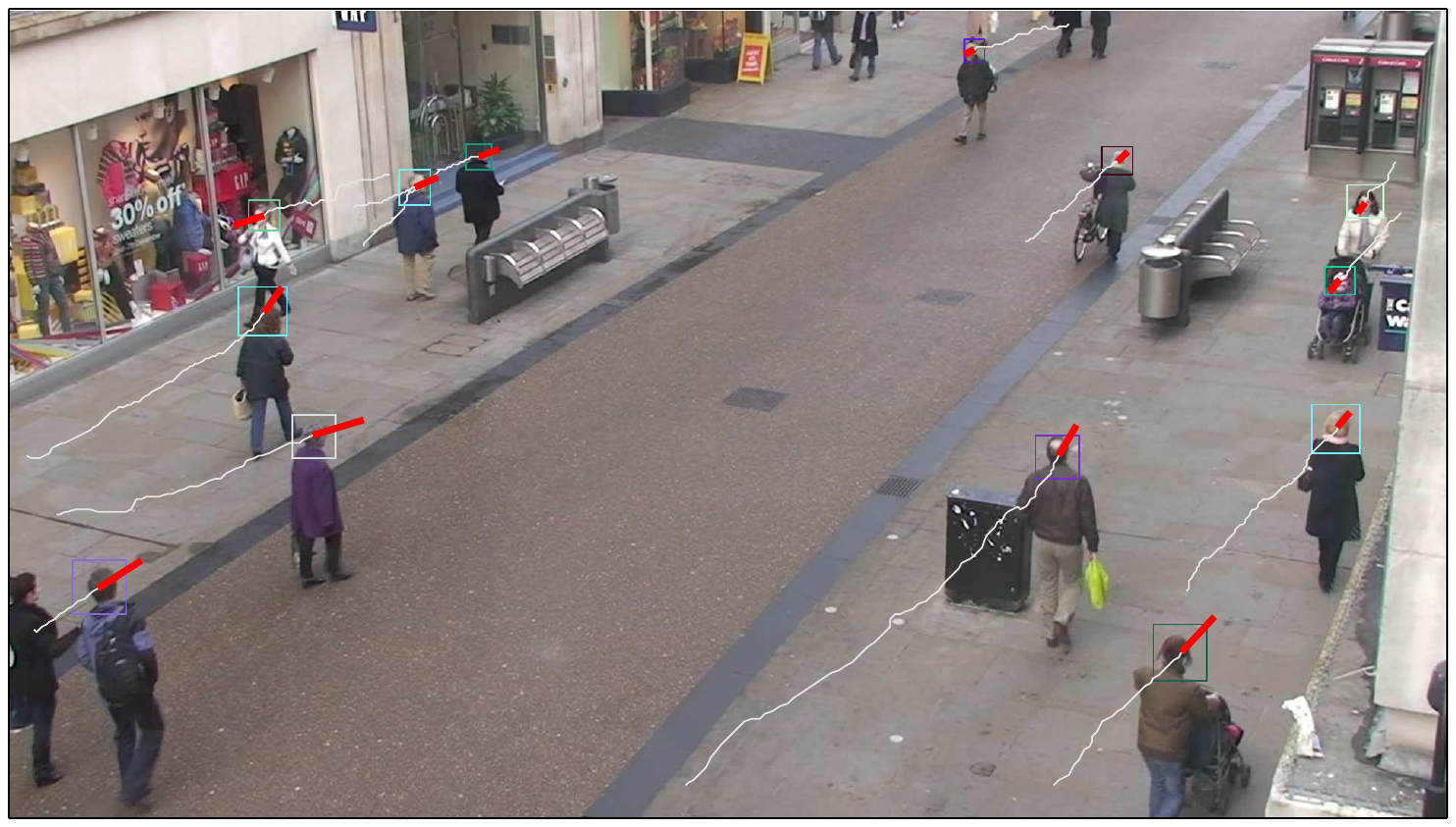}}\vspace{-5.5cm}
\centerline{\footnotesize (b)} \centerline{}
 \caption{ Frame 320 from the dataset used in performance evaluation: (a)
 Head and body detections (yellow and cyan rectangles, respectively); (b) tracker output  (the result of recursive
 processing of 79 previous frames and the current frame); track trajectories indicated by white lines;
 red lines are velocity vectors; squares indicate the heads of pedestrians}
 \label{f:1}
\end{figure*}

In this paper we adopt the ``tracking-by-detection'' approach to
pedestrian tracking, which has become very popular in computer
vision due to its applicability to moving un-calibrated cameras
\cite{breitenstein_09,benfold_11}. Typically pedestrian detections
are obtained using the Histogram of Oriented Gradients (HOG)
detector \cite{hog_05}, trained using either head images (for head
detections) or body images (for body detections). An example of head
and body detections is shown in Fig.\ref{f:1}.(a). Head and body
detections are unreliable in the sense that: (1) not all pedestrians
are detected in every frame; (2) the chance of false detections is
quite real, with the spatial density of false detections typically
non-uniform. This is evident in Fig.\ref{f:1}.(a).

If the state vector of each object (pedestrian) contains the
position (e.g. the head centroid), but not the size of the object,
then both the head and body detections are instances of {\em
imprecise} measurements: they represent rectangular regions
(two-dimensional intervals) within which the true object is possibly
located. As such, they can be modelled as random closed sets (rather
than random variables). The first tracking algorithm considered in
the paper is designed specifically for imprecise measurements: it
represents a multi-object tracker built from the Bernoulli filter
described in \cite{gning_ristic_mihaylova_tsp}. We will refer to
this tracker as Algorithm 1.

An imprecise measurement (e.g. an interval in the measurement space)
can always be converted to a precise but random measurement (e.g. a
point in the measurement space which is affected by additive noise).
The two remaining algorithms considered in the paper assume precise
random detections of heads/bodies for pedestrian tracking. Algorithm
2 is the same as Algorithm 1, but using the precise random
(Gaussian) measurements model. Algorithm 3 is based on the
Cardinalised PHD filter with data association \cite{petetin_11}.

Evaluation of multi-object tracking performance has been one of the
main stumbling blocks in advancing the scientific field of target
tracking. A large number of evaluation measures have been proposed,
both in the general context  (e.g. \cite{drummond_91,
colegrove_perf}) and specifically for video surveillance (e.g.
\cite{pingali_96, ellis_02, Bashir06, clear_mot}). At present there
is no consensus in sight on the preferred common approach. In this
paper tracking error will be measured using the recently formulated
Optimal sub-pattern assignment (OSPA) metric for tracks (or OSPA-T)
\cite{ospat}.  The OSPA-T metric has an important advantage over all
above mentioned performance metrics: it is a mathematically rigorous
metric (it satisfies the axioms of a metric) for measuring the
distance between two sets of tracks (i.e. between the ground truth
and the tracker output). OSPA-T  is also consistent with intuition,
as discussed in \cite{SchuhmacherOSPA07}.

The remainder of the paper is organised as follows. Section
\ref{s:2} describes the performance evaluation framework: the video
dataset, the method of pre-processing detections and the OSPA-T
metric. Section \ref{s:3} reviews the three random-set based
tracking algorithms. Section \ref{s:4} presents the experimental
results under various conditions, while the conclusions of this
study are drawn in Section \ref{s:5}.

\section{Performance evaluation framework}
\label{s:2}

\subsection{Video dataset and detections}
\label{s:11}
 The video dataset and the (hand labelled) ground truth
are downloaded from the website \cite{dataset}. The dataset is a
video of an occasionally busy town centre street (the average number
of pedestrians visible at any time is 16). The video was recorded in
high definition ($1920\times 1080$ pixels at $25$ fps). Only the
first 4500 frames of the video are used in the performance
evaluation. Frame number 320 of the dataset is shown in
Fig.\ref{f:1}.

Head detection and pedestrian body detection algorithms were applied
to each frame in the video sequence.  The fastHOG GPU
implementation~\cite{prisacariu_tr} of the Histogram of Oriented
Gradients (HOG) detection algorithm~\cite{hog_05} was used for both
detectors.  The HOG detector applies a sliding window to the image
at a regular grid of locations and scales, and classifies each
sub-window as containing or not containing an object (head or
pedestrian).  Classification is performed using a pre-trained linear
Support Vector Machine, the input to which is a set of block-wise
histograms of image gradient orientation. A classification threshold
of 0.75 was used for both detectors.  Sliding window detectors tend
to give multiple detections for one object due to their tolerance to
shift and scale, so a post-processing step groups overlapping
detections.

The head and pedestrian (whole body) detections have some
complementary characteristics.  The detector is only partially
tolerant to occlusions, so the head detector tends to have a higher
probability of detection since heads are generally more visible than
whole bodies in surveillance video.  However pedestrian textures are
more distinctive than head textures, so the head detector tends to
have a higher false alarm rate, picking up on round-ish background
objects such as clocks and signs.  The pedestrian detector is more
able to detect people at a distance where the head becomes too small
in the image.

Head and body detections are treated as if they are independent.
Each tracker can then be regarded as a centralised multi-source
fusion node, where one source of detections is the head detector
while the other is the body detector. The rectangles corresponding
to body detections are converted to head-like detections as follows.
Suppose a body detection rectangle is specified by its upper-left
corner $(\chi_b,\eta_b)$, width $w_b$ and height $h_b$. Then for its
corresponding head-like detection, the upper-left corner coordinates
are computed as: $\chi'_h = \chi_b + 0.325\,w_b$ and $\eta'_h =
\eta_b + 0.09 \, h_b$, while the width and height are $w'_h = 0.35
\, w_b$ and $h'_h = 0.19 \,h_b$, respectively.

\subsection{OSPA-T metric}
\label{s:ospat}

Traditional multi-object tracking performance measures describe
various aspects of tracking performance, such as {\em timeliness}
(e.g. track initiation delay, track overshoot), {\em track accuracy}
(e.g. position, heading, velocity error), {\em track continuity}
(e.g. track fragmentation, track labelling swaps) and  {\em false
tracks} (their count and duration). These measures are based on
heuristic, and it is unclear how to combine them into a single score
because they are correlated.

OSPA-T \cite{ospat} is defined as a theoretically rigorous distance
measure on the space of finite sets of tracks, and it has been
proven that it satisfies the axioms of a metric. The computation of
OSPA-T is described in Table \ref{aa:1}. Suppose we are given two
sets of tracks, the ground truth tracks $\{X^{(1)},\dots,X^{(L)}\}$
and estimated tracks $\{Y^{(1)},\dots,Y^{(R)}\}$. A track $X^{(l)}$,
$l=1,\dots,L$,  is defined as a temporal sequence
$X^{(l)}=(X_1^{(l)},\dots,X_K^{(l)})$ where each $X^{(l)}_k$,
$k=1,\dots,K$, is either an empty set (if track does not exist at
time $k$) or a singleton whose element is $(l,\xb_k)$. Here
$l\in\mathbb{N}$ is the track label and $\xb_k$ is its state at time
$k$. The labels of ground truth tracks are by convention adopted to
be $1,2,\dots,L$.

The first step in the computation of OSPA-T is to label the
estimated tracks (steps 3,4 and 5 in Table \ref{aa:1}). This first
involves finding the best assignment $\lambda^*$ of $R$ estimated
tracks to $L$ ground truth tracks. An assignment is a mapping
$\lambda^*(\ell)\in \emptyset \cup \{1,\dots,R\}$, for
$\ell=1,\dots,L$. This is typically carried out using a
two-dimensional assignment algorithm, such as the auction or Munkres
algorithm \cite{blackman_popoli_99}. If for an estimated track
$r=1,\dots,R$  exists a true track $\ell$ such that
$\lambda^*(\ell)=r$, then  track $r$ is assigned label $\ell$.
Estimated tracks which remain unassigned according to $\lambda^*$
are given labels different from all true track labels (i.e. integers
greater than $L$).

 Then, for each time step
$k=1,\dots,K$, the OSPA distance between the two labeled sets:
\begin{eqnarray}
\frak{X}_k & = &
   \{(l_1,\xb_{k,1}),\dots,(l_m,\xb_{k,m})\}\\
\frak{Y}_k & = &
   \{(s_1,\yb_{k,1}),\dots,(s_n,\yb_{k,n})\}
\end{eqnarray}
is computed. The set $\frak{X}_k$ represents the set of existing
ground truth labeled track states at time $k$; similarly
$\frak{Y}_k$ is the set of existing estimated labeled track states
at time $k$. The OSPA distance between these two labeled sets is
computed as \cite{ospat}:
\begin{align}
D&(\frak{X}_k,\frak{Y}_k) =  \nonumber \\
 & \left[\frac{1}{n}\left(\min_{\pi\in \Pi_n}\sum_{i=1}^m
\Big(d_{c}\big(\xbt_{k,i},\ybt_{k,\pi(i)}\big)\Big)^p + (n-m)\cdot
c^p\right)\right]^{1/p}  \label{e:ospa-t}
\end{align}
where $\xbt_{k,i}\equiv(l_i,\xb_{k,i})$,
$\ybt_{k,\pi(i)}\equiv(s_{\pi(i)},\yb_{k,\pi(i)})$ and
\begin{itemize}
\item $d_c(\xbt,\ybt) = \min(c,d(\xbt,\ybt))$ is the {\em cut-off distance} between
two tracks at $t_k$,  with $c>0$ being the cut-off parameter;
\item $d(\xbt,\ybt)$ is the {\em base distance}  between
two tracks at $t_k$;
\item $\Pi_n$ represents the set of permutations of length $m$ with
elements taken from $\{1,2,\dots,n\}$;
\item $p\in[1, \infty)$ is the OSPA metric order parameter.
\end{itemize}
For the case $m > n$, the definition is
$D_{p,c}(\frak{X}_,\frak{Y})=D_{p,c}(\frak{Y}_k,\frak{X}_k)$. If
both $\frak{X}_k$ and $\frak{Y}_k$ are empty sets (i.e. $m=n=0$),
the distance is zero.

\begin{table}[htb]
  \caption{Computation Steps of OSPA-T Metric}
\vspace{0.1cm}\hrule\vspace{0.1cm}
\begin{algorithmic}[1]{\small\renewcommand{\baselinestretch}{1.}
\Function{OSPA-T}{$\{X^{(1)},\dots,X^{(L)}\},\{Y^{(1)},\dots,Y^{(R)}\}$}
   \State $\%$ Label the estimated tracks
   \State For $j=1,\dots,R$, Label$[Y^{(j)}] = I$ (where $I>L$)
   \State Find $\lambda^*$, the globally best assignment of tracks
   $\{X^{(1)},\dots,X^{(L)}\}$ to $\{Y^{(1)},\dots,Y^{(R)}\}$
   \State For $i=1,\dots,L$, Label$[Y^{(\lambda^*(i))}]=
   \mbox{Label}[X^{(i)}]$
   \State $\%$ Compute the distance
   \State For $k=1,\dots,K$
   \State \hspace{.5cm} Form the labeled sets at $t_k$:\\
   \hspace{1.3cm}-Ground truth: $\frak{X}_k =
   \{(l_1,\xb_{k,1}),\dots,(l_m,\xb_{k,m})\}$\\
   \hspace{1.3cm}-Estimated: $\frak{Y}_k =
   \{(s_1,\yb_{k,1}),\dots,(s_n,\yb_{k,n})\}$
   \State \hspace{.5cm} Compute the OSPA distance between
   $\frak{X}_k$ and $\frak{Y}_k$
\EndFunction }\end{algorithmic} \vspace{0.1cm}\hrule\vspace{0.1cm}
\label{aa:1}
\end{table}

The base distance $d(\xbt,\ybt)$ is defined as:
\begin{equation}
d (\xbt,\ybt)  = \Big( d_\ell(\xb,\yb)^{p'} + d_\alpha(l,s)^{p'}
\Big)^{1/p'}, \label{e:base2}
\end{equation}
where:
 $p'\in[1, \infty)$ is the base distance order parameter;
 $d_\ell(\xb,\yb)$ is the localisation base distance, typically adopted as the
$p'$-norm: $d_\ell(\xb,\yb) = \| \xb-\yb \|_{p'}$;
 $d_\alpha(l,s)$ is the labeling error, adopted as:
$d_\alpha(s,t) = \alpha\,\bar{\delta}[s,t]$, where
$\bar{\delta}[i,j]$ is the complement of the Kroneker delta, that is
$\bar{\delta}[i,j]=0$ if $i=j$, and $\bar{\delta}[i,j]=1$ otherwise.
Parameter $\alpha\in[0,c]$ here controls the penalty assigned to the
labeling error $d(s,t)$ interpreted relative to the localisation
distance $d_\ell(\xb,\yb)$. The case $\alpha=0$ assigns no penalty,
and $\alpha=c$ assigns the maximum penalty.

Since in this paper we consider a sequence of a large number of
frames ($K=4500$), the OSPA-T is applied over non-overlapping
segments (blocks) of frames\footnote{The MATLAB source code for
computation of OSPA-T metric, including the head and body detections
for running and comparing different tracking algorithms, can be
obtained upon request from the first author.}.

\subsection{Base distance is a metric}

The base distance $d(\xbt,\ybt)$, defined in (\ref{e:base2}),
satisfies the three axioms of a metric: identity, symmetry and
triangle inequality. To prove identity and symmetry is trivial. The
proof of triangle inequality, presented in \cite[Sec.III.A]{ospat},
is wrong and this section presents the correct proof.

Let $\xbt=\left(l,\mathbf{x}\right)$,
$\ybt=\left(s,\mathbf{y}\right)$, $\zbt=\left(u,\mathbf{z}\right)$.
The following proof for the triangle inequality is given  in
\cite[Sec III.A]{ospat}
\begin{equation}
d\left(\mathbf{\tilde{x}},\mathbf{\tilde{y}}\right)^{p}\leq
d\left(\mathbf{\tilde{x}},\mathbf{\tilde{z}}\right)^{p}+d\left(\mathbf{\tilde{z}},\mathbf{\tilde{y}}\right)^{p}\label{eq:metric1}
\end{equation}
where in Sec.\ref{s:ospat} and \cite{ospat} notation $p'$ was used
instead of $p$. Equation (\ref{eq:metric1}) is wrong and this can be
seen for example by adopting: $p=2$,
$\mathbf{\tilde{x}}=\left(1,0\right)$,
$\mathbf{\tilde{y}}=\left(1,5\right)$,
$\mathbf{\tilde{z}}=\left(1,4.99\right)$. Then
$d\left(\mathbf{\tilde{x}},\mathbf{\tilde{y}}\right)^{p}=25$ and
$d\left(\mathbf{\tilde{x}},\mathbf{\tilde{z}}\right)^{p}+d\left(\mathbf{\tilde{z}},\mathbf{\tilde{y}}\right)^{p}\approx24.90$.
Moreover, (\ref{eq:metric1}) does not prove the triangle inequality.

We want to prove that
\begin{equation}
d\left(\xbt,\ybt\right)\leq
d\left(\xbt,\zbt\right)+d\left(\zbt,\ybt\right)
\end{equation}
where according to (\ref{e:base2})
\begin{equation}
d\left(\xbt,\ybt\right)^{p}=d_\ell\left(\mathbf{x},\mathbf{y}\right)^{p}+\alpha^{p}\overline{\delta}
\left[l,s\right]. \label{eq:metric}
\end{equation}

As $d_\ell\left(\cdot,\cdot\right)$ is a metric, it meets the
triangle inequality 
\begin{equation}
d_{\ell}\left(\mathbf{x},\mathbf{y}\right)\leq
d_{\ell}\left(\mathbf{x},\mathbf{z}\right)+d_{\ell}\left(\mathbf{z},\mathbf{y}\right)
\end{equation}
As both sides of the inequality are positive numbers and $p\geq1$
\begin{equation}
d_{\ell}\left(\mathbf{x},\mathbf{y}\right)^{p}\leq\left(d_{\ell}\left(\mathbf{x},\mathbf{z}\right)+d_{\ell}\left(\mathbf{z},\mathbf{y}\right)\right)^{p}\label{eq:rp1}
\end{equation}
We also have that
\begin{equation}
\alpha\overline{\delta}\left[l,s\right]\leq\alpha\overline{\delta}
\left[l,u\right]+\alpha\overline{\delta}\left[u,s\right]\label{e:ineq77}
\end{equation}
As both sides of inequality (\ref{e:ineq77}) are positive and
$p\geq1$
\begin{equation}
\left(\alpha\overline{\delta}\left[l,s\right]\right)^{p}\leq
\left(\alpha\overline{\delta}\left[l,u\right]+\alpha\overline{\delta}\left[u,s\right]\right)^{p}\label{eq:rp2}
\end{equation}
Using (\ref{eq:rp1}) and (\ref{eq:rp2})
\begin{align}
d_{\ell}&\left(\mathbf{x},\mathbf{y}\right)^{p}+\left(\alpha\overline{\delta}\left[l,s\right]\right)^{p}\leq\nonumber
\\
&\bigg(d_{\ell}\left(\xb,\zb\right)+d_{b}\left(\mathbf{z},\mathbf{y}\right)\bigg)^{p}+
\bigg(\alpha\overline{\delta}\left[l,u\right]+\alpha\overline{\delta}\left[u,s\right]\bigg)^{p}
\end{align}
that is
\begin{align}
&\sqrt[p]{d_{\ell}\left(\xb,\yb\right)^{p}+\left(\alpha\overline{\delta}\left[l,s\right]\right)^{p}}
\leq\nonumber\\
&\sqrt[p]{\left(d_{\ell}\left(\xb,\zb\right)+d_{\ell}\left(\zb,\yb\right)\right)^{p}+
\left(\alpha\overline{\delta}\left[l,u\right]+\alpha\overline{\delta}\left[u,s\right]\right)^{p}}
\label{eq:rp3}
\end{align}
As $p\geq1$, using the Minkowski inequality \cite{Hardy_book34} on
the right hand side of (\ref{eq:rp3})
\begin{align}
&\sqrt[p]{\left(d_{\ell}\left(\xb,\zb\right)+d_{\ell}\left(\zb,\yb\right)\right)^{p}+
\left(\alpha\overline{\delta}\left[l,u\right]+\alpha\overline{\delta}\left[u,s\right]\right)^{p}}
\leq
\nonumber\\
&
\sqrt[p]{d_{\ell}\left(\xb,\zb\right)^{p}+\alpha^{p}\overline{\delta}\left[l,u\right]}+
\sqrt[p]{d_{\ell}\left(\mathbf{z},\mathbf{y}\right)^{p}+\alpha^{p}\overline{\delta}\left[u,s\right]}\label{eq:rp4}
\end{align}
Finally, using (\ref{eq:rp3}) and (\ref{eq:rp4}), we get
\begin{align}
&\sqrt[p]{d_{\ell}\left(\mathbf{x},\mathbf{y}\right)^{p}+\alpha^{p}\overline{\delta}\left[l,s\right]}
\leq \nonumber
\\
&
\sqrt[p]{d_{\ell}\left(\mathbf{x},\mathbf{z}\right)^{p}+\alpha^{p}\overline{\delta}\left[l,u\right]}+
\sqrt[p]{d_{\ell}\left(\mathbf{z},\mathbf{y}\right)^{p}+\alpha^{p}\overline{\delta}\left[u,s\right]}
\end{align}
The proof is finished using (\ref{eq:metric}):
\begin{equation}
d\left(\mathbf{\tilde{x}},\mathbf{\tilde{y}}\right)\leq
d\left(\mathbf{\tilde{x}},\mathbf{\tilde{z}}\right)+d\left(\mathbf{\tilde{z}},\mathbf{\tilde{y}}\right).
\end{equation}

%

\section{Description of Algorithms}
\label{s:3}

The state vector of a single object is adopted for all algorithms as
 $\xb = \left[\begin{matrix}x&\dot{x}& y
&\dot{y}\end{matrix}\right]^\intercal$, where $(x,y)$ is the
position (in pixels) of the pedestrian head centroid and
$(\dot{x},\dot{y})$ is its velocity vector (in pixels/s). The number
of objects from frame to frame varies. The random finite set of {\em
head} detections at frame $k$ is denoted $Z_k^{(1)}$. Accordingly,
the random set of {\em head-like} body detections (see the
explanation in the last sentence of Sec.\ref{s:11}) is $Z_k^{(2)}$.

Algorithms 1 and 2 are based on the multi-sensor Bernoulli filter
\cite{BTVo_2011}, where the ``sensors'' are the two types of
pedestrian head detections. Separate and independent Bernoulli
filters are run for each target.  Target interactions are taken care
of by the appropriately increased  clutter intensity, as  in
\cite{musicki_lm_08}. This multi-object tracking algorithm has been
described in some detail in \cite{Fus10_ospat}.  The difference
between Algorithms 1 and 2 is in the model of the single-object
likelihood function. Let $\zetaB\in Z_k^{(i)}$, for $i=1,2$, be a
detection resulting from an object (i.e. a pedestrian head) in the
state $\xb$. A head detection is a rectangle, thus $\zetaB$ is
specified by a tuple $(\chi,\eta,w,h)$, where $(\chi,\eta)$
determines its upper-left corner, while $w$ and $h$ are the width
and height, respectively.

The single-object likelihood function used in Algorithm 1 treats the
detection $\zetaB$ as an imprecise measurement and is defined as in
\cite{gning_ristic_mihaylova_tsp}:
\begin{equation}
g_k^{(i)}(\zeta|\xb) =
\varphi(\mathbf{H}\xb;\underline{\zeta},\mathbf{\Sigma}^{(i)}) -
\varphi(\mathbf{H}\xb;\overline{\zeta},\mathbf{\Sigma}^{(i)})
\label{e:glf}
\end{equation}
where $\varphi(\zb;\mu,\mathbf{\Sigma}^{(i)})$ is the Gaussian
cumulative distribution function with mean $\mu$ and covariance
$\mathbf{\Sigma}^{(i)} =\mbox{diag}[{\sigma_x^{(i)}}^2,\;
{\sigma_y^{(i)}}^2]$; $\underline{\zeta}$ and $\overline{\zeta}$ are
the lower and upper bound of the rectangle, and
$\mathbf{H}=\left[\begin{matrix}1&0&
0&0\\0&0&1&0\end{matrix}\right]$. If
$\sigma_x^{(i)}=\sigma_y^{(i)}=0$, then (\ref{e:glf}) simply states
that $g_k^{(i)}(\zeta|\xb) = 1$ if $(x,y)$ is inside the rectangle
$\zeta$, and zero otherwise. The algorithm is applied to the video
dataset using $\sigma_x^{(1)}=1$ and $\sigma_x^{(2)}=25$.

Algorithms 2 and 3 first convert the rectangular detection $\zetaB$
into a point measurement $\zb = \left[\chi+w/2,\;
\eta+h/2\right]^\intercal$, with the associated covariance matrix
$\mathbf{R} = \mbox{diag}[(w/6)^2,\; (h/6)^2]$. Then the
single-object likelihood function of $\zb$ is adopted as:
\begin{equation}
g_k^{(i)}(\zb|\xb) = \mathcal{N}(\zb; \mathbf{H}\xb, \mathbf{R})
\label{e:likfunk}
\end{equation}
where  $\cal{N}( \mathbf{m};\mu,\mathbf{P})$ is the Gaussian
probability density function with mean $\mu$ and covariance
$\mathbf{P}$.

Algorithm 3 is based on the Cardinalised PHD (CPHD) filter
\cite{mahler_cphdf_07}, but with additional logic to deal with track
labeling. The key idea of \cite{petetin_11} is to form the clusters
of targets, and to apply the CPHD filter update  to each cluster
separately. The update uses every available detection (measurement)
to calculate the {\em weight} of the track-to-measurement
association. The weight of no-measurement association is also
computed. Finally these weights are used to form an association
matrix which is solved using a two-dimensional assignment algorithm
(e.g. auction, Munkres). At last each predicted track is updated
with the measurement which has been assigned to it by the assignment
algorithm. Since we have at our disposal two types of detections
($Z_k^{(1)}$ are head detections, and $Z_k^{(2)}$ are head-like body
detections), the update step in Algorithm 3 is applied twice, first
using $Z_k^{(1)}$ and then using $Z_k^{(2)}$. Although this is not
an optimal approach \cite{mahler_poisson_magic1}, it has been
suggested as a reasonable
approximation. 

\begin{table}[htb]
 \caption{A Summary of the Contesting Tracking Algorithms}
\begin{center}
\begin{tabular}{ccc}\hline\hline
Alg.  & Likelihood function & Method  \\
\hline
1.  &  Eq.(\ref{e:glf}) &  Multi-Bernoulli Tracker of \cite{Fus10_ospat}  \\
\hline 2. &   Eq.(\ref{e:likfunk}) & Multi-Bernoulli Tracker of
\cite{Fus10_ospat} \\ \hline 3. & Eq.(\ref{e:likfunk}) & CPHD based
tracker \cite{petetin_11}\\
 \hline\hline
\end{tabular}
\end{center}
\label{t:algs}
\end{table}

All three algorithms used the same clutter maps (one map for heads,
the other for body detections). The probability of detection was set
to $P_D^{(1)}=0.58$ and $P_D^{(2)}=0.52$. A short summary of
algorithms is given in Table \ref{t:algs}.

\section{Numerical results}
\label{s:4}

\begin{figure*}[tbh]
\vspace{-7cm}\centerline{\includegraphics[width=20.cm]{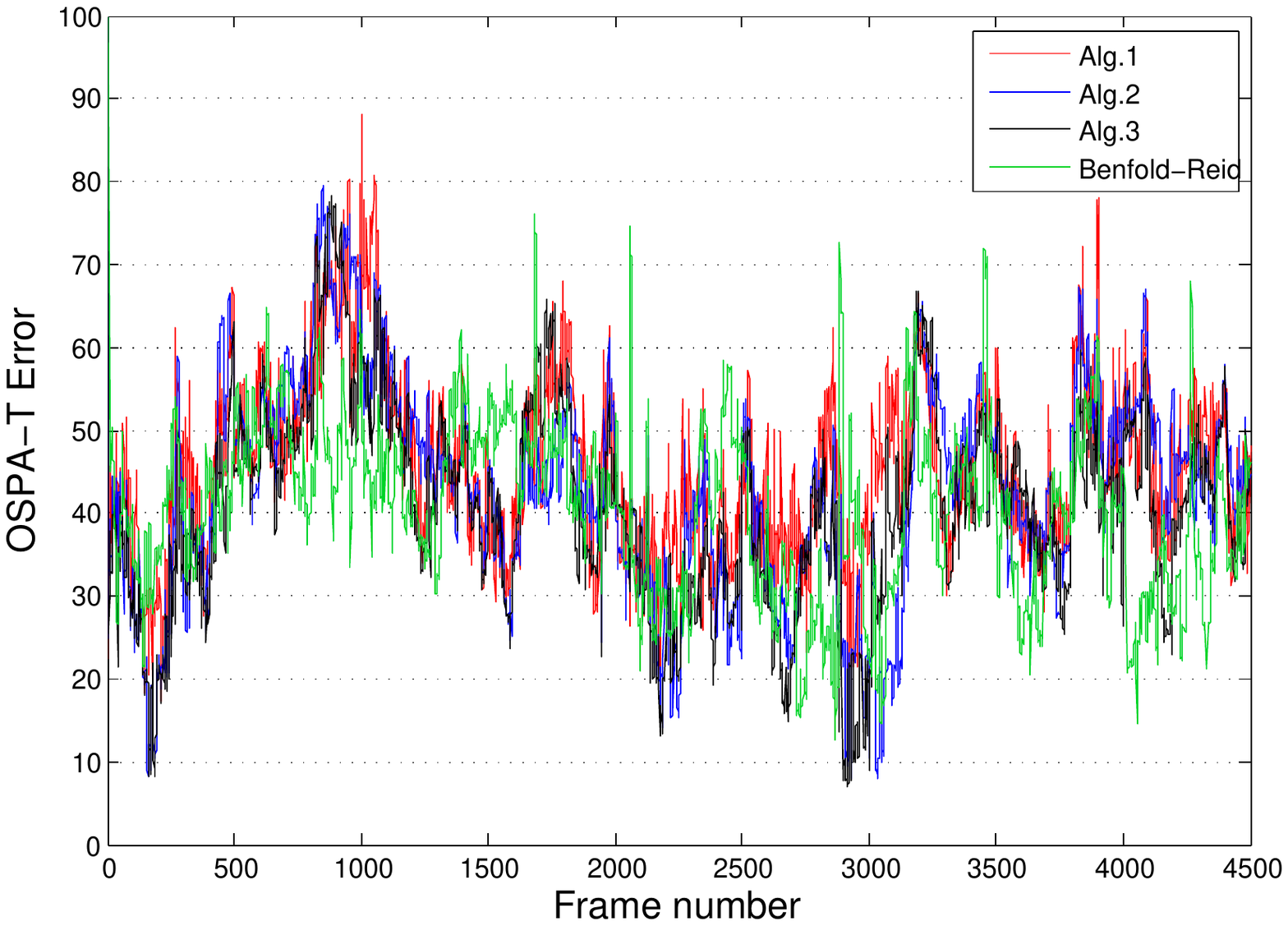}}
\vspace{-7cm}
 \caption{Comparison of tracking algorithms: OSPA-T error for the sequence of $4500$ frames; OSPA-T
metric parameters: $p=p'=1$, $c=100$ and $\alpha = 75$}
 \label{f:ospat}
\end{figure*}

The localisation base distance of the OSPA-T error $d(\xb,\yb) = \|
\xb-\yb \|_{p'}$ only takes into account the positional error (i.e.
neglecting the velocity error). Fig.\ref{f:ospat} shows the
resulting OSPA-T error for the three random-set based tracking
algorithms. The parameters of the OSPA-T metric used in evaluation:
$p=p'=1$, $c=100$ and $\alpha = 75$. Identical head detections and
body-to-head converted detections, from every frame, have been used
by all three algorithms. Fig.\ref{f:ospat} also shows, as a
guideline, the OSPA-T error of the Benfold-Reid (BR) algorithm
\cite{benfold_11}, whose tracking results are available online
\cite{dataset}. We point out that the comparison between the BR
algorithm and the three random-set based trackers is not fair
because the BR algorithm is a smoother (operates as a batch
algorithm over a sliding window of image frames) and does not use
body/head detections in every frame. From Fig.\ref{f:ospat} one can
observe that ranking of the algorithms according to OSPA-T varies
with time. For example, from frame number 800 to 1100, the BR
algorithm is far superior than the random-set based trackers, but
the opposite is true from frame 1400 to 1600. In order to obtain an
overall ranking, the time averaged OSPA-T error has be computed: its
value for  Algorithms 1, 2, 3 and the BR algorithm  is 45.2, 42.8,
40.7 and 40.4, respectively. The conclusion is that the most
accurate of the three random-set tracking algorithms is Algorithm 3.
Furthermore, it appears that the imprecise measurement model is not
justified in the adopted context: the transformation of head and
body-to-head rectangles (imprecise detections) into random precise
measurement points provides better tracking results. This can be
explained by the nature of head and body-to-head rectangular
detections; it has been observed that if a detection is not false,
then its rectangular centre is a very accurate estimate of the
centre of a pedestrian head. Thus the likelihood (\ref{e:glf}),
which is based on the interpretation that the true head centroid is
somewhere inside the rectangle, appears to be too cautious and
consequently does not use the full information content of a
measurement.

We repeated the OSPA-T error computations for $\alpha=0$ (no penalty
for the labeling error). This case corresponds to the original OSPA
error proposed in \cite{SchuhmacherOSPA07}.  The obtained time
averaged OSPA-T error for Algorithms 1, 2, 3 and the Benfold-Reid
algorithm in this case were 34.1,   29.5,   27.4, and 30.2,
respectively. Again Algorithm 3 performs the best among the
random-set based trackers, and even outperforms the Benfold-Reid
algorithm.  This result reveals that the major problem with
Algorithm 3 is the lack of track consistency (too many broken
tracks), which by adopting $\alpha=0$ is not penalised. Track
consistency can be improved by smoothing over multiple image frames
(to be considered in the future work).

\begin{figure*}[bth]
\vspace{-7cm}\centerline{\includegraphics[width=20cm]{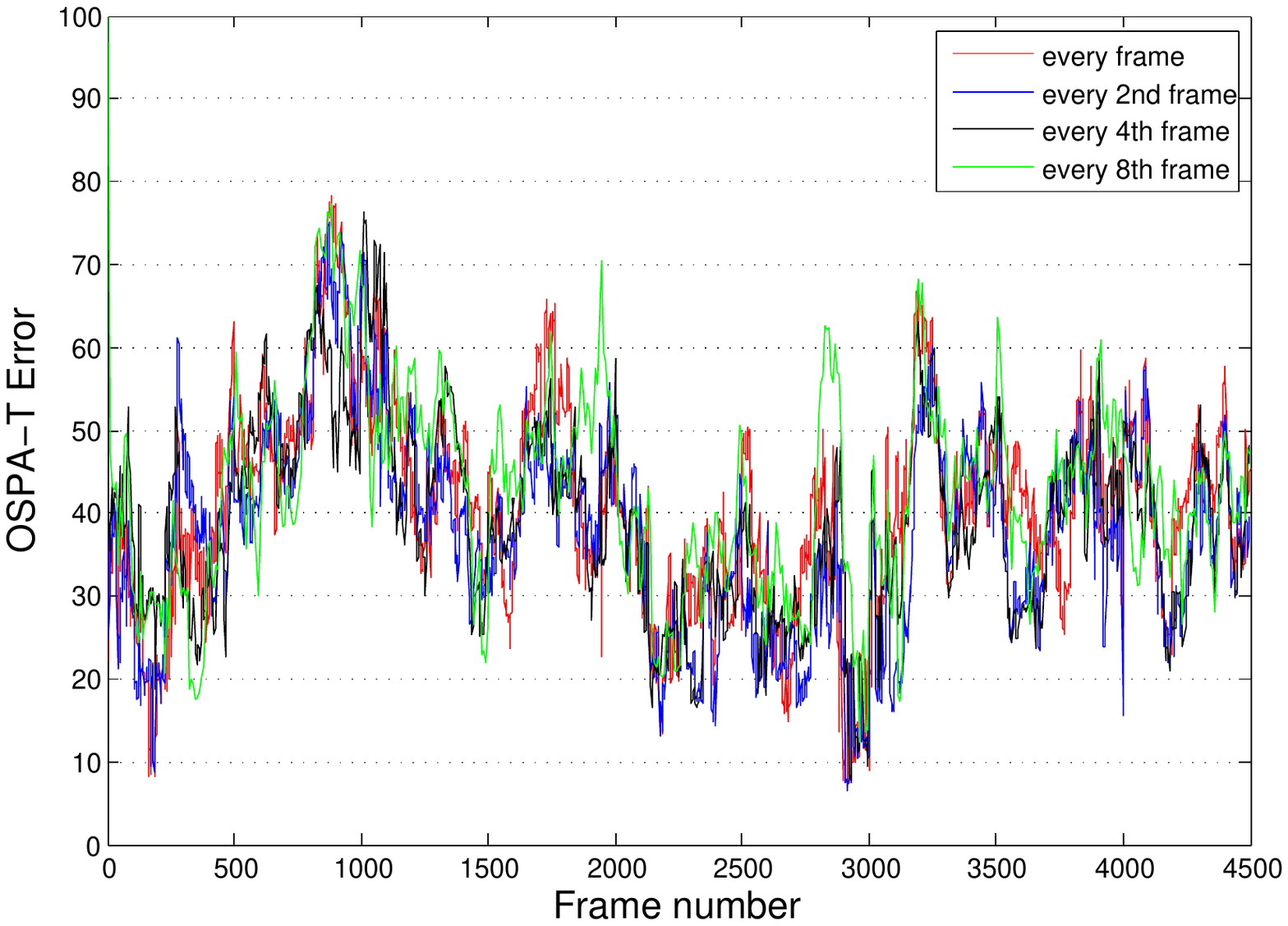}}
\vspace{-7cm}
 \caption{The influence of detection frequency on Algorithm 3: OSPA-T error for the sequence of $4500$ frames; OSPA-T
metric parameters: $p=p'=1$, $c=100$ and $\alpha = 75$. }
 \label{f:3}
\end{figure*}

Head and body detection algorithms are very computationally
intensive and consequently in real-time applications it may not be
possible to provide them at every image frame. Next we compare the
OSPA-T error performance of Algorithm 3 for the situations where
head and body detections are available for: (1) every frame, (2)
every 2nd frame, (3) every 4th frame and (4) every 8th frame. The
results are shown in Fig.\ref{f:3}. We note that the error
performance does not change dramatically with the reduced frequency
of head and body detections.  The  time averaged OSPA-T error for
the four cases are:  40.7,  37.9, 39.2, and 42.7. Somewhat
surprisingly, using body/head detections every 2nd and every 4th
frame, reduces the number of false tracks and overall improves the
accuracy. Only when body/head detections become available only every
8th frame, some of the true tracks start to be missing occasionally
and consequently the OSPA-T error performance deteriorates.


\section{Conclusions}
\label{s:5}

The paper presented a framework for performance evaluation of
multi-object trackers. The framework is illustrated in the context
of video tracking by comparison of three random-set based pedestrian
tracking algorithms, using a video data set of a busy town centre.
The multi-object tracking error was evaluated using the ``OSPA for
tracks'' (OSPA-T) metric. The OSPA-T metric has an important
property that it satisfies the axioms of a metric. The mathematical
proof the triangle inequality axiom is presented in the paper.

The results of performance evaluation indicate that the CPHD based
tracker of \cite{petetin_11} performs the best. Although this is a
single-frame recursive algorithm, its performance is comparable to
that of \cite{benfold_11} (which operates over a sliding window of
frames). Future work will consider a smoothing version of the
algorithm in \cite{petetin_11} since a delay by a few frames in
reporting the tracks is tolerable and has the potential to further
improve the tracking accuracy.

\bibliographystyle{plain}

\end{document}